\documentclass[10pt]{article}
\usepackage[utf8]{inputenc}
\usepackage[T1]{fontenc}
\usepackage{amsmath}
\usepackage{amsthm}
\usepackage{amsfonts}
\usepackage{amssymb}
\usepackage{makeidx}
\usepackage{graphicx}
\usepackage{algorithm2e}
\usepackage{lmodern}
\usepackage{mathtools}
\usepackage{url}
\usepackage{xcolor}
\usepackage[left=2cm,right=2cm,top=1cm,bottom=2cm]{geometry}
\usepackage{multirow}
\usepackage{caption}
\usepackage{subcaption}
\usepackage{tabularx} 
\usepackage{booktabs}

\author{Alfred Kofi Adzika    \thanks{Ohio University OH, Athens. 
		\href{mailto:aa832423@ohio.edu}
{aa832423@ohio.edu}} 
  \and   Prudence Djagba\thanks{ Michigan State University
		\href{mailto:arame.sow@aims-senegal.org}{ djagbapr@msu.edu}}
  }
\title{Inference with k-means}

\usepackage{hyperref}
\usepackage{cleveref}

\newtheorem{thm}{Theorem}

\theoremstyle{remark}

\theoremstyle{definition}
\newtheorem{defn}[thm]{Definition}
\date{}
\begin{document}
\maketitle	
\begin{abstract} 
\noindent

This thesis aims to invent new approaches for making inferences with the k-means algorithm. k-means is an iterative clustering algorithm which starts by randomly assigning k centroids, then assigns data points to the nearest centroid, and updates centroids based on the mean of assigned points. This process continues until convergence, forming k clusters where each point belongs to the nearest centroid.  This research investigates the prediction of the last component of data points obtained from a distribution of clustered data using the online balanced k-means approach. Through extensive experimentation and analysis, key findings have emerged. It is observed that a larger number of clusters or partitions tends to yield lower errors while increasing the number of assigned data points does not significantly improve inference errors. Reducing losses in the learning process does not significantly impact overall inference errors. Indicating that as learning is going on inference errors remains unchanged.  Recommendations include the need for specialized inference techniques to better estimate data points derived from multi-clustered data and exploring methods that yield improved results with larger assigned datasets. By addressing these recommendations, this research advances the accuracy and reliability of inferences made with the k-means algorithm, bridging the gap between clustering and non-parametric density estimation and inference.
 \\


\end{abstract}
 \section{Introduction}

 Over the years, many research paths and discipline find their roots in natural phenomena. For instance, Physics, Oceanography, Ecology, Astrophysics, and Neuroscience are all vast academic disciplines that stem from the physical occurrence around humanity. Similarly, machine learning is not isolated from such areas of research. Machine learning is a field that seeks to understand physical phenomena with the goal of making predictions at desired precision. 

In data science, clustering is a helpful tool. It is a technique for discovering cluster structure in a dataset that is distinguished by the highest degree of similarity inside a cluster and the highest degree of dissimilarity across clusters. The first clustering technique employed by biologists and social scientists was hierarchical clustering, whereas cluster analysis evolved into a subfield of statistical multivariate analysis \cite{jain1988algorithms}, \cite{kaufman2009finding}. In terms of machine learning, it belongs to unsupervised learning. 
Clustering techniques can be broadly categorized as nonparametric approaches and probabilistic model-based approaches from a statistical perspective. The mixture likelihood approach to clustering is utilized \cite{mclachlan1988mixture} because the probability model-based approaches assume that the data points come from a mixture probability model. The expectation and maximization (EM) algorithm is most frequently utilized in model-based techniques \cite{dempster1977maximum}, \cite{yu2018convergence}.

Clustering techniques for nonparametric approaches can be separated into hierarchical and partitional techniques, with partitional techniques being the most utilized \cite{kaufman2009finding}, \cite{jian2009data}, \cite{yang2018fully}. These techniques are based primarily on an objective function of similarity or dissimilarity measures.

Partitional approaches generally assume that the data set may be represented by finite cluster prototypes with independent objective functions. Therefore, it is crucial for partition algorithms to provide the dissimilarity (or distance) between a point and a cluster prototype. The k-means algorithm is acknowledged as the most established and widely used partitioning technique \cite{jain1988algorithms}, \cite{macqueen1967some}. k-means clustering has received a lot of attention in the literature and has been used in a wide range of relevant fields \cite{alhawarat2018revisiting}, \cite{zhu2019efficient} acknowledged.

Delving into the target of the thesis, the goal of this work is to explore inference methods associated with the online balanced k-means algorithm which is a variant of the k-means algorithm, and experiment for their precision. This is an uncommon path to thread however, inspiration to conduct this work sprung from \cite{nedergaard2022k}. 

Our thesis is organized as follows; the first chapter provides an introduction to the thesis, followed by a section dedicated to the review of related articles. Next, a background is provided to solidify the foundations necessary to ensure a fruitful study of this research. Chapter four is a vivid explanation of the approaches employed in conducting this research. The next chapter entails a breakdown and explanation of the results. Lastly, the final chapter comprises limitations, recommendations, and a concluding note.

\section{Background} 

The Background chapter provides a comprehensive overview of the key concepts and methodologies that form the foundation of the project. This chapter explores the fundamental principles of k-means clustering algorithms, highlighting their widespread usage and significance in data analysis and density estimation. Additionally, the chapter delves into the concept of Non-parametric density estimation, discussing its importance in accurately estimating the underlying distributions of data. The exploration of Voronoi diagrams and Voronoi density estimation techniques provide valuable insights into spatial partitioning and density estimation approaches that serve as crucial components for making inferences with the k-means algorithm. By examining and understanding these essential concepts, we lay the groundwork for developing novel approaches in this research, aiming to advance the capabilities of the k-means algorithm in clustering and inference tasks. The Background chapter sets the stage for the subsequent chapters, offering a comprehensive understanding of the relevant theories and techniques necessary to comprehend the advancements made in the project.

\subsection{k-Means clustering}

k-means is an unsupervised machine learning technique that partitions data points into k clusters. The technique operates by iteratively assigning n data points $ (x_1, x_2,\ldots, x_n)  $ to the nearest mean and updating the means to the centroids of the points given to each cluster $ (c_1, c_2,\ldots, c_n)  $ after randomly initializing k points, referred to as means or cluster centroids. The algorithm keeps running until the means ceases to evolve or a predetermined number of iterations has been reached. The algorithm is presented below in two steps that are iterated until convergence.

\begin{enumerate}
	\item Each data point is assigned to cluster with the closest mean:
	\begin{equation}
		x \in c_i \Longleftrightarrow \forall j\ne i. \quad d(x,\mu_i) \le d(x,\mu_j).
	\end{equation}

 \item  Cluster mean is updated to the mean of the newly formed cluster:
  \begin{equation}
  	\mu_i \longleftarrow \frac{1}{n_i}\sum_{x \in c_i}x
  \end{equation}
where $ n_i  =  |c_i|$ is the number of points in cluster $ c_i $.
\end{enumerate}
k-means aims to minimize the sum of squared distances between data points and their assigned centroid, known as the Within-Cluster Sum of Squares (WCSS). It is sensitive to initial centroid placement and converges to local minima. 
There are other variants of the k-means algorithm which is relevant to this research. Namely, online k-means, the balanced k-means, and the online balanced k-means which is a combination of the online k-means, and the balanced k-means.

\begin{enumerate}
	\item \textbf{Online k-means}
	
	Online k-means is a clustering algorithm that can be used to cluster data that is streaming in real-time. The algorithm works by iteratively assigning new data points to the cluster with the nearest centroid. The centroids of the clusters are then updated to reflect the new data points. Online k-means is useful when data is received one by one or in chunks and allows for updating the model as new data is received. However, it is dependent on the order in which the data is received \cite{pmlr-v201-bhattacharjee23b}.
	\begin{enumerate}
		\item A data point is assigned to cluster with the closest mean:
		\begin{equation}
			x \in c_i \Longleftrightarrow \forall j\ne i. \quad 	d(x,\mu_i) \le d(x,\mu_j).
		\end{equation}
		
		\item  Cluster means are updated to the mean of the newly formed cluster:
		\begin{equation}
			\mu_i \longleftarrow \alpha x + (1-\alpha)\mu_i
		\end{equation}
		where the hyperparameter $ \alpha \in (0,1) $ is the learning rate.
	\end{enumerate}
	\item \textbf{Balanced k-means}
	
	Balanced k-means is a variation of the k-means clustering algorithm that aims to produce clusters of equal size. In balanced k-means, the cluster sizes are constrained to be equal, and the algorithm optimizes the mean square error for given cluster sizes \cite{nedergaard2022k}.
	\begin{enumerate}
		\item A data point is assigned to cluster:
		\begin{equation}
			x \in c_i \Longleftrightarrow \forall j\ne i. \quad	d(x,\mu_i)-w_i \le d(x,\mu_j)-w_j.
		\end{equation}
		
		\item  Cluster mean is updated:
		\begin{equation}
			\mu_i \longleftarrow \frac{1}{n_i}\sum_{x \in c_i}x
		\end{equation}
	
	or
		\begin{equation}
			\mu_i \longleftarrow \alpha x + (1-\alpha)\mu_i.
		\end{equation}
	\item Weights are updated
	\begin{align*}
            n_i &\longleftarrow n_i + 1\\
		w_i &= \beta \dfrac{n_i -\mathbb{E}[n]}{\mathbb{V}[n]}
	\end{align*}

where $ \mathbb{E}[n] $ and $ \mathbb{V}[n] $ represents the mean and variance of the number of points in the clusters and $ \beta \in \mathbb{R} $ is the hyperparameter known as the balancing scale.
	
	\end{enumerate}
 \item \textbf{Online balanced k-means}

 This is another variant of k-means that merges both the online and balanced k-means variants. Just like the online variant k-means, the online balanced k-means receives each data point in turns. In addition, the online balanced k-means penalizes clusters with lots of points. Hence, the clusters created are of similar sizes.
 


\end{enumerate}

\subsection{Non-parametric density estimation}

Non-parametric density estimation is a statistical method used to estimate the probability density function of a random variable without assuming any specific form for the function. In non-parametric density estimation, the objective is to estimate a probability density function fX from a set of points D. Non-parametric density estimation methods include histogram estimator, naive estimator, kernel density estimator (KDE), K-nearest neighborhood (KNN) estimator, and Voronoi density estimation.



Kernel density estimation involves placing a kernel function at each data point and summing them to estimate the density. The choice of kernel function and bandwidth parameters can significantly impact the accuracy of the estimation. Selecting an appropriate bandwidth can be challenging, as too narrow or too wide a bandwidth can lead to underfitting or oversmoothing, respectively \cite{wkeglarczyk2018kernel}.

Histogram-based methods divide the data range into bins and count the number of data points falling into each bin. The width and number of bins need to be carefully chosen to capture the underlying distribution accurately. However, histograms may suffer from discontinuities at bin boundaries and are sensitive to binning choices \cite{silverman2018density}.

Nearest-neighbor methods estimate the density by considering the density around each data point based on the distance to its neighboring points. These methods can be computationally intensive, especially with large datasets, as they require pairwise distance calculations \cite{ferraty2006nonparametric}.

Evaluating the performance of non-parametric density estimation techniques can also be challenging. Metrics such as cross-validation or information criteria can be used, but they may not always provide straightforward guidance for selecting the best method or parameter values.

Even though \cite{cadre2006kernel} suggests that Non-Parametric density estimation is a very powerful tool in unsupervised learning, Polianskii et. al. revealed in \cite{polianskii2022voronoi} that density estimation methods, such as the kernel density estimator and histograms, have a common limitation of being biased towards a fixed local geometry. This means that the estimates made by the kernel density estimator near a sample are influenced by the level sets of the chosen kernel, which are ellipsoids of high estimated probability. Similarly, histograms suffer from a bias towards the geometry of the cells of the tessellation, on which the estimated probability density function is constant. In other words, these methods are limited in their ability to capture the true underlying distribution of the data and may produce biased estimates due to their dependence on the local geometry of the data.

As a remedy to the mentioned limitations, Ord in 1978 \cite{ord1978many} suggests that the Voronoi density estimation dealt with the challenges discussed earlier.

\subsection{Voronoi diagrams} 
A Voronoi diagram divides a set of points in a plane into an equal number of cells so that each point, $ p_i $, is contained within a cell $ c_i $ made up of all areas that are closer to $ p_i $ than to any other point in P. 

The relation between the Voronoi diagram and k-means lies in how they conceptually represent the clustering process. At each iteration of the k-means algorithm, the centroids are updated based on the mean of the data points assigned to each cluster. This means that the centroid represents the mean or average position of all the data points assigned to that cluster.

Considering this, you can interpret each cluster centroid in the k-means algorithm as a point in the Voronoi diagram. The Voronoi regions associated with each centroid represent the data points that are closest to that centroid. In other words, the Voronoi regions correspond to the assignment of data points to different clusters in the k-means algorithm.

By visualizing the Voronoi diagram associated with the centroids obtained from the k-means algorithm, you can see how the data space is divided into distinct regions, each representing a cluster. This visualization can provide insights into the structure of the data and how the algorithm has grouped similar points together.

In summary, the Voronoi diagram helps illustrate the assignment of data points to clusters in the k-means algorithm, providing a geometric representation of the clustering process. Below is an example of the Voronoi diagram with black points representing the centroids of the partitions. These centroids also qualify to be the means of clusters of k-means clustering.  

\begin{figure}[h!]
    \centering
    \includegraphics[scale = 0.2]{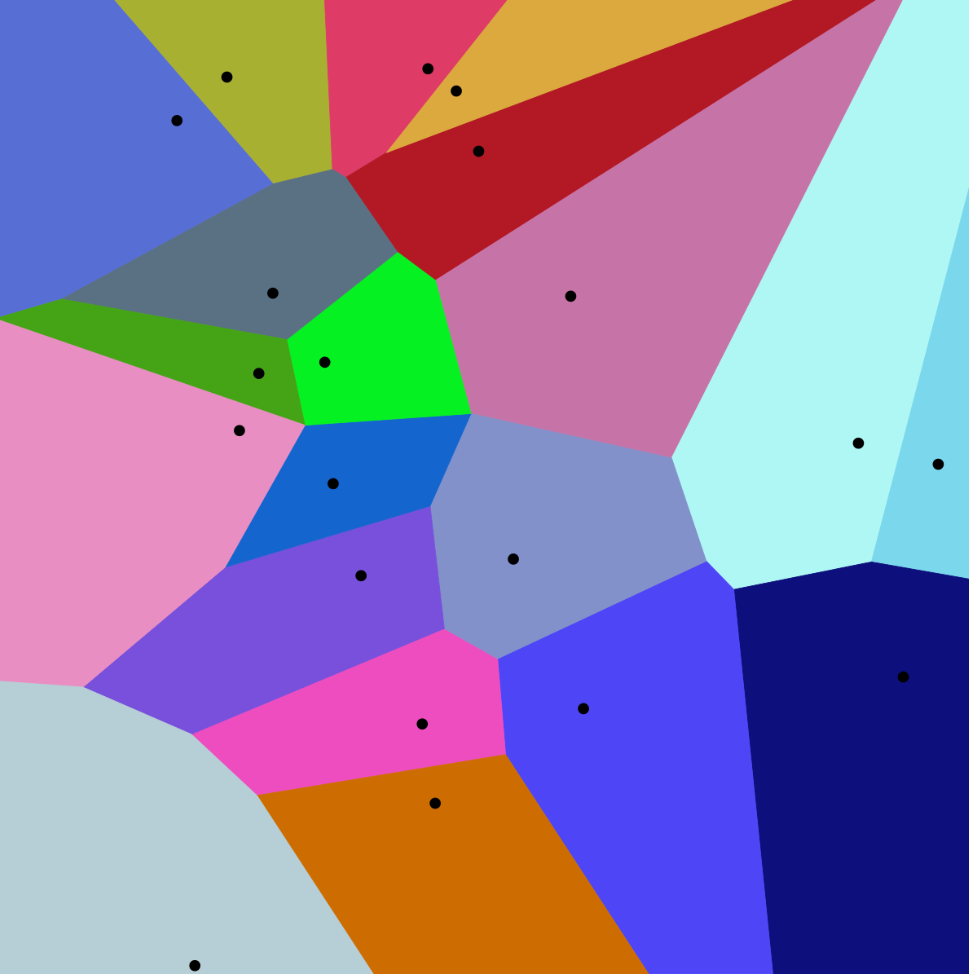}
    \caption{A Voronoi diagram}
    \label{fig:vd}
\end{figure}

\subsection{Voronoi density estimation}
\begin{defn}
	Let P be a set of points. The Voronoi cell of a point $ p_i \in P $ is defined as $$ c_i = {x \in X \mid \forall p_j \in P \quad d(x,p_i) \le d(x,p_j)} $$ where $ i \ne j $.
\end{defn}
The ambient space X is covered by the Voronoi cells in a pairwise disjoint fashion. Thus the cells intersect at their boundaries. The collection of all the cells are called Voronoi tessellations generated by the set of points in this case P.
\begin{defn}
	For a point $ x \in X $ the Voronoi Density Estimator (VDE) is defined almost everywhere as $$ p(x) = \frac{1}{|P|Vol(c(x))} $$ where |$\cdot$| denotes cardinality and $Vol$ denotes a finite Borel measure.
\end{defn}

The function p(x) represents a locally constant probability density function (PDF) on the space X, which in turn defines a probability measure called $Vol$. Under this distribution, each cell in the space is equally likely. Additionally, the normalization of $Vol$ within each cell aligns with the restriction placed on it. In other words, the probability measure $Vol$ is appropriately adjusted and normalized within each individual cell based on the local constant PDF defined by $ p(x)$. Overall, Voronoi density estimation provides a way to estimate the density of a distribution by partitioning the data space into Voronoi cells and considering locally constant PDFs within each cell \cite{polianskii2022voronoi}.


	
	
	


\section{Methods}%
\subsection{Inference with k-means}
Inference with k-means is the act of predicting unknown components of a data point. This data point is picked from a data set that has the same distribution as the point. Literature has not presented any guidelines on how inferences could be done with k-means hence, the researcher had to brainstorm to invent new ones. Seven novel inference approaches were discovered. These inference methods will be well expounded on in subsequent chapters.

\subsection{The online balanced k-means} 
 The implementation of the online balanced k-means involves a combination of the online k-means and the balanced k-means. To do this, the researcher went through available notes by Alexander Nedergaard. Also, Richard Duda provided an algorithm \cite{richd2007} in the setup of the online k-means which aided in the implementation of the online balanced k-means. The following algorithm was used in building the online balanced k-means:

 \begin{enumerate}
     \item Assign point to a cluster:
     $$
     x \in c_i \Longleftrightarrow \forall j \ne i. D(x, \mu_i ) - w_i \le D( x, \mu_j) - w_j.
     $$
     \item Update centers:
     \begin{align*}
     n_i \gets & n_i +1\\
         \mu_i \gets & \alpha x + (1-\alpha)\mu_i.
     \end{align*}
     \item Update cluster weights
     $
    w_i = \beta \dfrac{n_i -\mathbb{E}[n]}{\mathbb{V}[n]}.
     $
     \item Return cluster index $(i)$.
     \label{enu obk_agorithm}
 \end{enumerate}
 
\subsection{Data generation approaches}
This section outlines the diverse data generation approaches utilized to obtain a wide range of data instances, allowing us to assess the performance variations of our inference methods effectively.
Data was generated using random data-generating functions in Python. The distributions used include:
\begin{enumerate}
	\item Uniform Distribution $ X \sim U(-1, 1) $.
	
	$f(x) = \frac{1}{2}$ where $-1 \le x \le 1$
	\item Normal distribution $ X \sim N(0,1)  $
	
	$ f(x) = \frac{1}{\sqrt{2\pi}} e^{-\frac{x^2}{2}} \text{ where } x \in \mathbb{R}$.
	\item Gamma Distribution
	$ X \sim \Gamma(1,1) $
	
	$ f(x) = e^{- x} \text{ where } x > 0 $.
	\item Special data-generating functions in Python such as make blobs and moon-shaped data.
\end{enumerate}

Also, the generated data was classified into three categories based on the number of clusters associated with each of them.

\subsection{Methods for generating data with different relations between them}
All the data was generated using the following steps:

Assume there are $ d $ dimensions.
\begin{enumerate}
	\item Generate data from a specific distribution for each of the $ d-1 $ features. These will form the input features.
	\item Operate a mathematical operation row-wise on each of the input features to obtain a value for the last feature (column).
	\item Return all the data in the columns based on the number of rows required by the user.
\end{enumerate}

\subsubsection{Data with one true cluster}

\begin{enumerate}
    \item Data generated from a uniform distribution includes:
    \begin{enumerate}
    	\item uniform: In generating this data, all of its features are randomly derived from a uniform distribution.
    	\item uniform squared: With this data generation approach, all features are derived from a uniform distribution with the exception of the final feature which is derived from the sum of squares of the entries corresponding to that column.  
    	\item uniform cube: Data generated with this approach requires that all features are generated from a uniform distribution, except for the final feature which is calculated as the sum of cubes of the entries corresponding to that column.
	\end{enumerate}
    \item Data generated from a normal distribution
    \begin{enumerate}
    	\item normal: This data is entirely generated from a normally distributed sample. 
    	\item normal squared: All of its features are obtained from a normal distribution excluding the last feature. The final feature is derived from the sum of squares of the entries corresponding to that column.
    	\item normal cube: With this data, only the last column is not randomly generated from a normal distribution. The last column is a sum of the cube of the entries in the preceding columns that correspond to the final column.
    \end{enumerate}
    \item Data generated from a gamma distribution distribution
    \begin{enumerate}
    	\item gamma: All features here are generated randomly from the Gamma distribution.
    	\item gamma squared: Its last column contains the square of data from its preceding columns. The data from its preceding column were derived randomly from the Gamma distribution.
    	\item gamma cube: The last column is a cube of the preceding columns.
    \end{enumerate}
\end{enumerate}

\subsubsection{Data with two true clusters}
\begin{enumerate}
    \item Data generated from two uniform distributions: With this data-generating approach, the first half of the data on each column is obtained randomly from a Uniform distribution ranging between -1 and 1 and the second half has a Uniform distribution ranging between 5 and 7.
    \item Data generated from two normal distributions: With this data-generating approach, the first half of the data on each column is obtained randomly from a Normal distribution with parameters 0 and 1 and the second half also has a Normal distribution with parameters 6 and 1.
    \item Data generated from two gamma distribution distributions: Here, data is generated from two separate gamma distributions.
\end{enumerate}
\subsection{Data with three true clusters}
\begin{enumerate}
    \item Data generated from three uniform distributions: This approach involves the generation of data from three different Uniform distributions.
    \item Data generated from three normal distributions: Data is generated from three normal distributions.
    \item Data generated from two gamma distribution distributions and a normal distribution. This method involves a blend of two gamma distributions and a normal distribution.
\end{enumerate}

\subsection{Inference methods with k-means}
This section is dedicated to the demonstration of how inference is done with the online balanced k-means. 

Each of the inference methods is targeted at predicting the last component of data points generated. That is, given $ \mathbf{x_i} $ to be $(x_0, x_1, x_2, \ldots, x_{d-2}, x_{d-1})_i  $, the aim is to estimate $ x_{d-1} $.

\subsubsection{The Euclidean distance approach}
This approach estimates the last component of a data point based on the closest centroid by measuring the Euclidean distance between the known components and all the centroids.
\begin{align*}
		\text{min distance index} &= \min_{\forall j}\{d(x_i, \mu_j)\}  \\
		\text{where } \mathbf{x_i} &= (x_0, x_1, x_2, \ldots, x_{d-2})_i   \\
		\mathbf{\mu_j} &= (\mu_0, \mu_1, \mu_2, \ldots, \mu_{d-2})_j\quad \text{ where }  j\in \mathbb{N}, \text{ and } j\le k \\
		x_{d-1} &= \mu_{d-1}.
\end{align*}

\subsubsection{The normalized weights approach }
The normalized weights approach estimates the last component of a data point using weights which are probability estimates depicting the closest with high probabilities. A mathematical representation of the weight is represented as follows. 

	\begin{align*}
		w_i &= \dfrac{e^{\beta_i \cdot\min_{ \forall j }{d(x_i, \mu_j)}}}{\sum_{\forall i} e^{\beta_i \cdot d(x_i, \mu_j)} }\\
	x_{d-1} &= \sum_{i = 1}^{k} w_i \cdot \mu_i.
	\end{align*}
\subsubsection{The cluster size approach} 
	This inference approach is similar to the normalized weight approach. However, the cluster size approach computes its weight differently. The weight is determined as follows:
	
	One has to first find the size of the five clusters closest to the point to be estimated. The sizes are to be used as weights. The following is the mathematical representation of the weights:
	
	Suppose $ \mu_0, \mu_1, \mu_2, \mu_3, \mu_4 $ are the closest centroids to the point $ x $ and the counts associated to their clusters are $ t_{0}, t_{1}, t_{2}, t_{3}, t_{4}$
	\begin{align*}
		w_j &=  \frac{t_{j}}{t_0+t_1+t_2+t_3+t_4} \\
		x_{d-1} &= \sum_{j=0}^4 w_{j} \cdot \mu_{j, d-1}.
	\end{align*}
\subsubsection{The overall mean and the normalized weights approach}
	In this case, we associate the estimated value with the overall mean of the data and the normalized weight approach. This method of estimating $ x_{d-1} $ is computed as follows:
	\begin{align*}
		x_{d-1} = \alpha \cdot \text{overall\_mean}(x) + (1- \alpha) \cdot (\text{norm\_weights}(x)).
	\end{align*}

\subsubsection{Merging the normalized weights and the cluster size approach } 
Similar to the previous approach, this inference method merges two different estimated values. It involves the normalized weights and the cluster size approach. It is computed as follows.
\begin{align*}
	x_{d-1} = \alpha \cdot (\text{norm\_weights}(x)) + (1- \alpha) \cdot \text{cluster\_size}(x).
\end{align*}
\subsubsection{Merging the normalized weights and the Euclidean distance approach }
This approach is a merge between the normalized weight approach and the Euclidean distance. It is computed as follows:
\begin{align*}
	x_{d-1} = \alpha \cdot (\text{norm\_weights}(x)) + (1- \alpha) \cdot \text{euclid\_dist}(x).
\end{align*}
\subsubsection{Cluster size with exponential weights approach}
This approach is an upgrade of the cluster weight approach. We introduce exponential and beta to the weights with the aim of refining the estimate better. The mathematical rendition is as follows:

Suppose $ \mu_0, \mu_1, \mu_2, \mu_3, \mu_4 $ are the closest to the point $ x $ and the counts associated to their clusters are $ t_{0}, t_{1}, t_{2}, t_{3}, t_{4} $.
 
\begin{align*}
	w_i =&  \exp{\left(\frac{-\beta \cdot t_{i}}{t_0 + t_1 + t_2 + t_3 + t_4 }\right)}\\
	x_{d-1} =& \sum_{i=0}^4 w_{i} \cdot \mu_{i, d-1}.
\end{align*}
\subsection{Errors and Losses}
\subsubsection{Compute the performance of each inference method}
	The performance of each inference method is determined by how well the inference method estimates $ x_{d-1} $, the last component of a point in the test data. Hence we use the squared distance approach to evaluate the total errors of a given inference method. The smaller the error the better the performance and vice versa. The errors are computed as follows:
	
	Let infer$(x)$ represent the inference models where $x$ is the data point to be inferred.
	$$ 
		\text{Infer\_error} = \sum_{\forall x} (x - \text{infer}(x)).
	$$ 

\subsubsection{Computing the loss associated with the generated data}
The loss of the online balanced k-means is computed as follows:
\begin{align*}
	\text{loss} = \sum_{\forall x} \min_{\forall i} d(x, \mu_i).
\end{align*}

Great amount of this thesis was done in python. We started from the implementation of the online balanced k-means, data generation approaches, inference methods with k-means, errors and losses computations and ended with the plotting of graphs. All these steps were implemented in the Python programming language in a Jupiter notebook. Kindly find attached the codes.\href{https://colab.research.google.com/drive/1SmpPj7oDEBjxBwLBGbNkx_Ss9BwY2qqE?usp=sharing}{\textcolor{black}{ Tap here to have access to the codes.}}

In conclusion, this methodology has provided a comprehensive framework for evaluating various inference methods with the k-means algorithm. We explored the traditional k-means approach and its extension, the online balanced k-means, both of which are crucial in clustering analysis. To ensure diverse and comprehensive testing, we devised multiple data generation approaches, encompassing features derived from varied distributions.

Additionally, the methodology sheds light on the computation of losses and inference errors, crucial for assessing the performance of the proposed methods. By carefully considering these metrics, we will analyze the inference methods to gain insights into the effectiveness of the inference techniques.

The combination of these components has allowed us to systematically analyze and compare the inference performance under various scenarios. This rigorous approach ensures the robustness and reliability of our results, providing a solid foundation for drawing meaningful conclusions from our experiments.

Moving forward, these insights will play a pivotal role in understanding the strengths and weaknesses of different inference methods.

\section{Results and discussions} 

In this section, we present the results and discussions of this project, which focuses on inventing new approaches to making inferences with the k-means algorithm. The chapter focuses on shedding light on the performance and potential of the proposed methods.  The Results and Discussions chapter showcases the contributions and insights gained toward advancing the field of clustering and inference techniques.
The following are procedures employed in this section.
\begin{enumerate}
    \item Partitioning of data into balanced clusters.
    \item Experimenting for the best hyperparameters.
    \item Investigating for the best-performing inference methods.
\end{enumerate}

\subsection{Visualising the online balanced k-means}
\begin{figure}[h!]
    \centering
    \caption{An illustration of the partitioning of the Gaussian and Uniform distribution using the online balanced k-means.}\includegraphics[scale=.4]{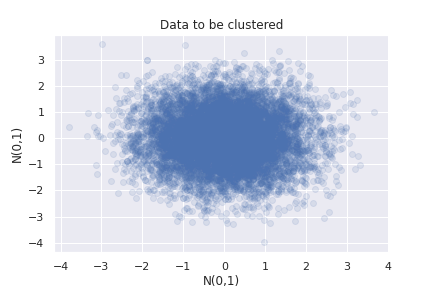} $  $
    \includegraphics[scale=.4]{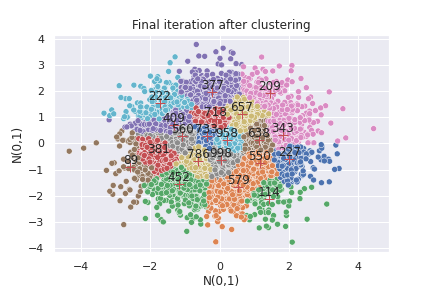}
    \\[\smallskipamount]
    \includegraphics[scale=.4]{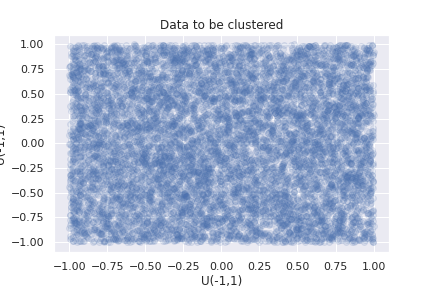} $  $
    \includegraphics[scale=.4]{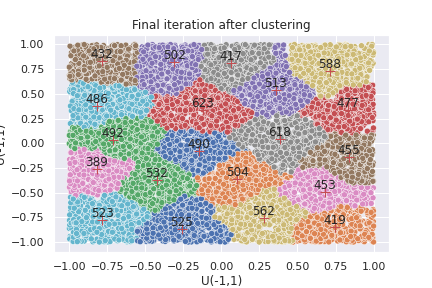}
    \label{fig:obk}
\end{figure}

\begin{figure}[h!]
    \centering
\includegraphics[width=\textwidth, height = 18cm]{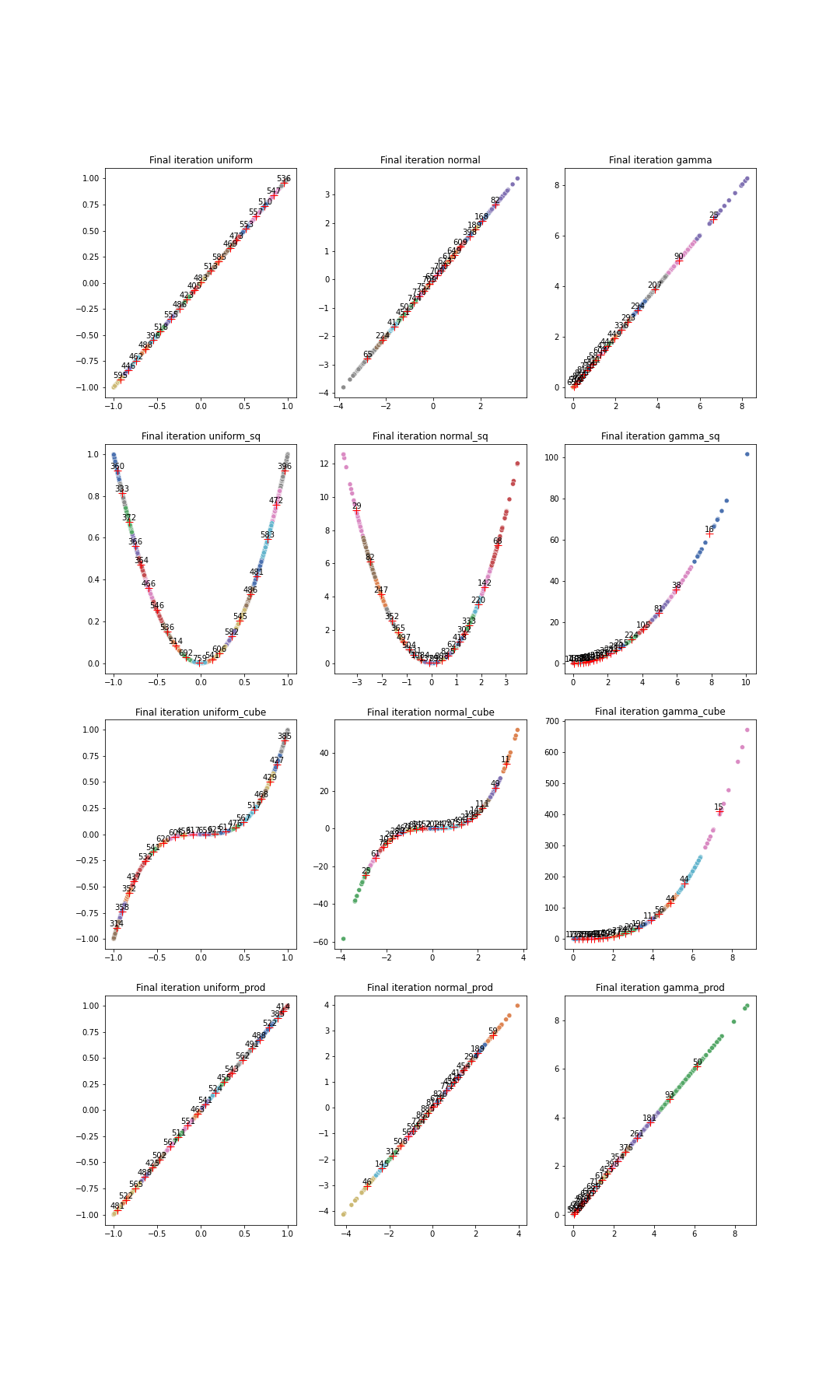}
    \caption{Data generated with different relations between them.}
    \label{fig:data_gen_a}
\end{figure}

\begin{figure}[h!]
    \centering
\includegraphics[width=\textwidth, height = 18cm]{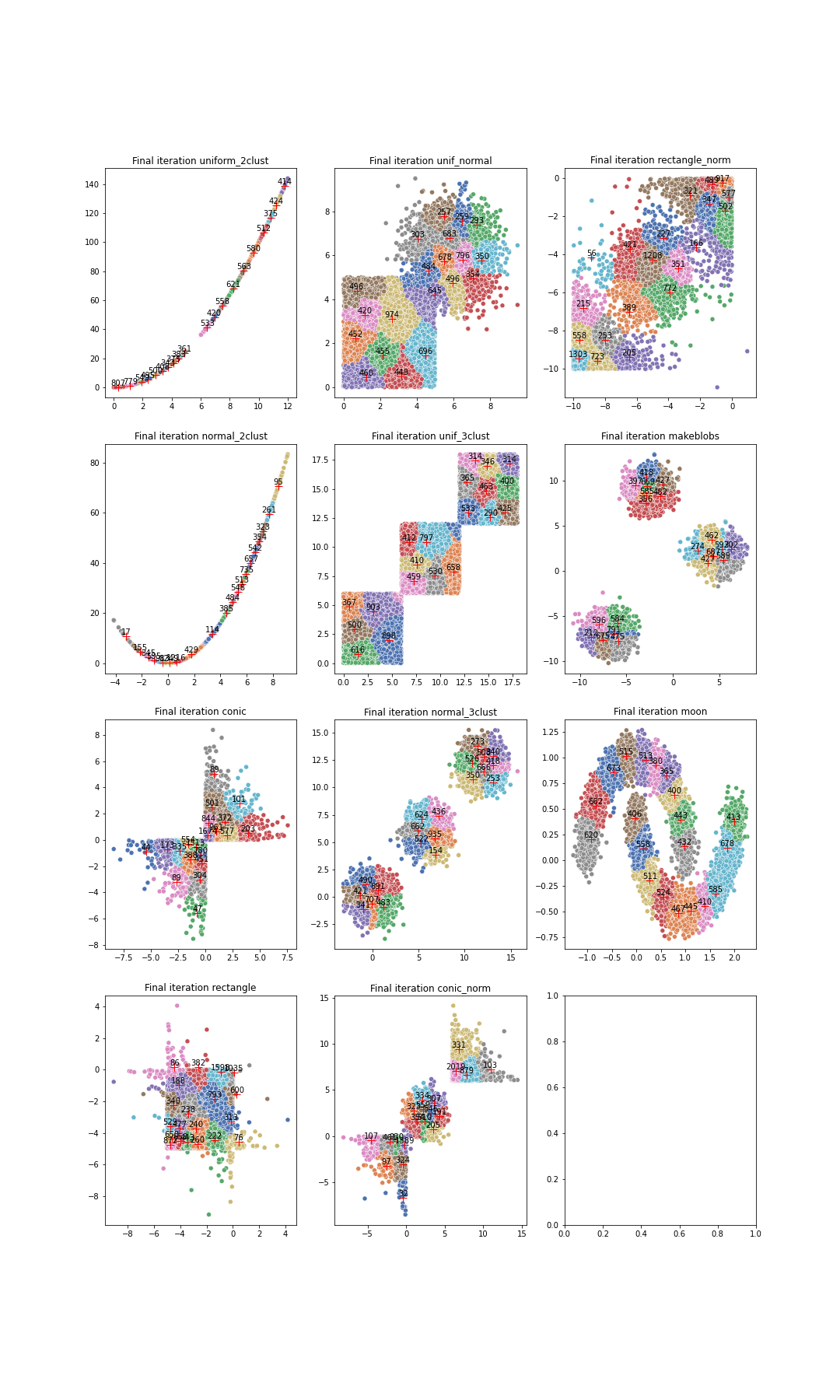}
    \caption{Data generated with different relations between them.}
    \label{fig:data_gen_b}
\end{figure}

Figure \ref{fig:obk} shows the partitions created by the online balanced k-means algorithm. To obtain these plots, one needs to follow the steps described in Section \ref{enu obk_agorithm}. This result provides us with data that would aid our inference methods. Furthermore, we observe that each cluster preserves information relating to the distribution of the data set. The diagrams demonstrate how the online balanced k-means transition a dataset into a compartmentalized dataset with balanced clusters. To visualize the partitions of the other generated data used in this work, refer to Figure \ref{fig:data_gen_a} and Figure \ref{fig:data_gen_b}.

\subsection{Performance of the inference methods over given hyperparameters}
\begin{figure}[h!]
    \centering
    \includegraphics[scale = 0.5]{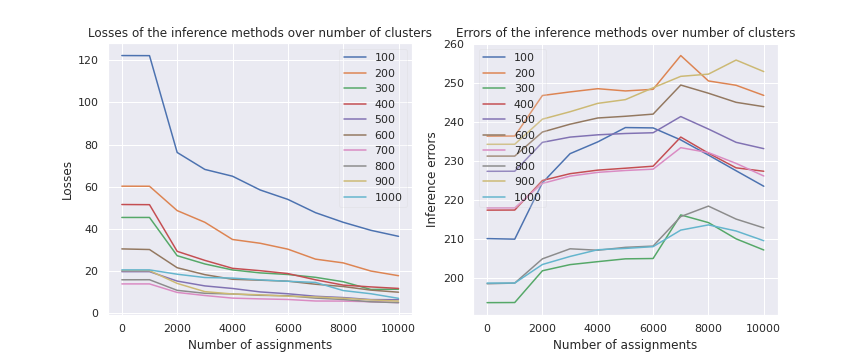}
    \caption{\textbf{Effects of the number of clusters on losses and errors over every 1000 iterations}}
    \label{fig: Loss_n_clust}
\end{figure}

\subsubsection{Number of clusters (k):} 

This crucial hyperparameter determines the number of clusters for the online balanced k-means algorithm, taking only positive values. Its significance lies in influencing the performance of inference methods. Figure \ref{fig: Loss_n_clust} illustrates how losses vary with 1000 data point assignments and presents this effect for 10 different cluster numbers, aiding k selection for accurate estimates. The left-hand graph depicts losses for k values between 100 and 1000, while the right-hand graph shows inference errors associated with k. The results indicate decreasing losses with increasing data point assignments, validating the learning process. Additionally, higher cluster numbers exhibit more pronounced convergence towards zero.

In the graph located to the left of Figure \ref{fig: Loss_n_clust}, we depict curves that illustrate the number of clusters varying from 100 to 1000. Notably, the inference errors exhibit a stabilizing pattern with each data point assignment. Among these different numbers of clusters, the most optimal choice appears to be 300, as it consistently yields the smallest inference errors.


\subsubsection{The learning rates ($\alpha$):}
\begin{figure}[h!]
    \centering
    \includegraphics[scale = 0.5]{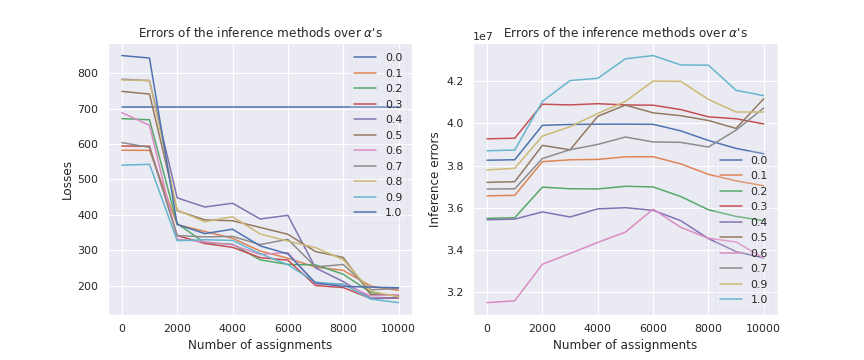}
    \caption{\textbf{Effects of the learning rates on losses and errors over every 1000 iterations}}
    \label{fig: Losses and errors associated with alpha}
\end{figure}

The learning rate ensures that the centroid is influenced by both its previous centroid ($\mu_j$) and the new data point ($x$). As more data points are assigned to the cluster, the influence of the new data point remains the same. 
The overall performance of the $\alpha$ is exhibited in Figure \ref{fig: Losses and errors associated with alpha}. The plot on the left-hand side shows how losses reduce after more data points are assigned. The plot on the right-hand shows how the errors remain stabilized as more data are assigned. 

From Figure \ref{fig: Losses and errors associated with alpha}, the best learning rate is $\alpha=$ 0.6, followed by $\alpha=$0.4 then $\alpha=$ 0.2. Also, learning rates closest to the value, one has proved to have the largest errors. For instance, the curves corresponding to $\alpha=$ 0.9 and $\alpha=$ 1 lay above all the curves. 


\subsubsection{The balance factor ($\beta$):}

\begin{figure}[h!]
    \centering
    \includegraphics[scale = 0.5]{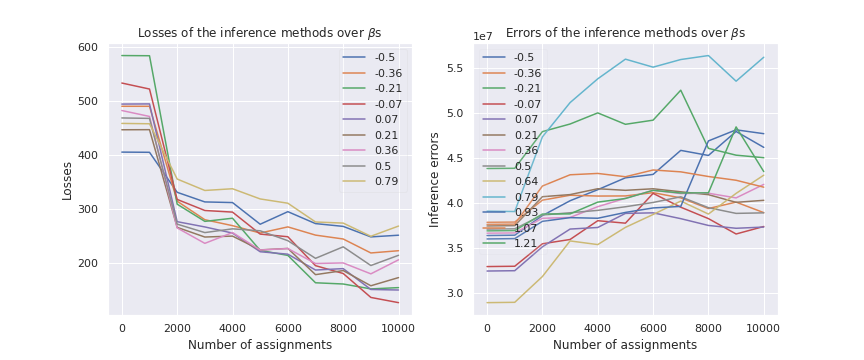}
    \caption{\textbf{Effects of the balance rate on losses and errors over every 1000 iterations}}
    \label{fig: Losses and errors associated with beta}
\end{figure}

The balance factor plays a crucial role in maintaining balanced cluster sizes after each assignment. As shown in Figure \ref{fig: Losses and errors associated with beta}, the behavior of the balance factor mirrors that of hyperparameters k and $\alpha$. Losses decrease following data point assignments, while errors steadily increase. Optimal balance rates, depicted in Figure \ref{fig: Losses and errors associated with beta}, lie within the range of -0.21 to 0.7, leading to the best performance in the clustering process.

\subsection{Investigating for the best-performing inference methods.}

\begin{figure}
    \includegraphics[width=.24\textwidth]{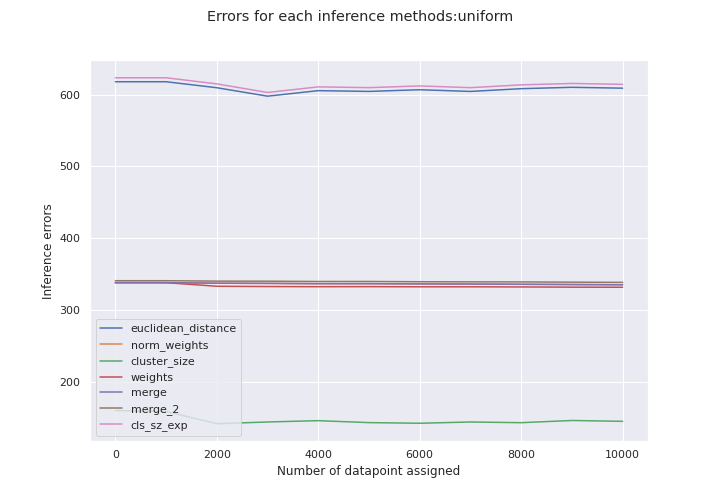}\hfill
    \includegraphics[width=.24\textwidth]{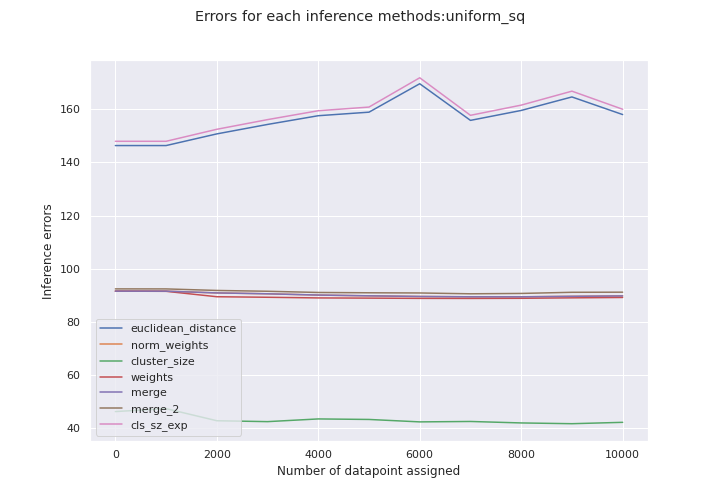}\hfill
    \includegraphics[width=.24\textwidth]{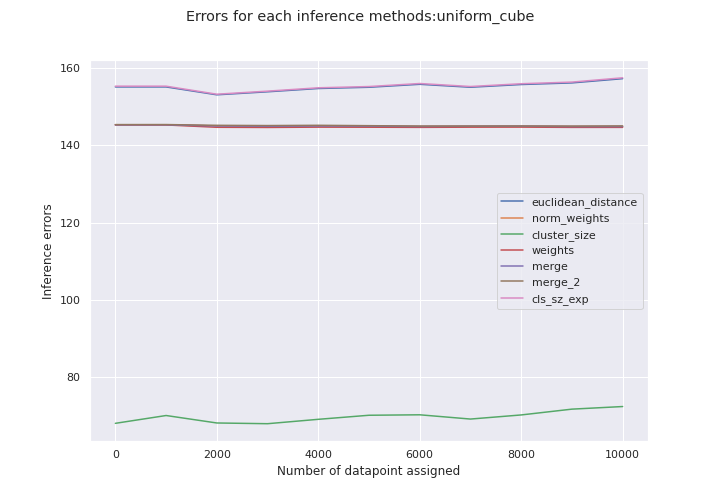}\hfill
    \includegraphics[width=.24\textwidth]{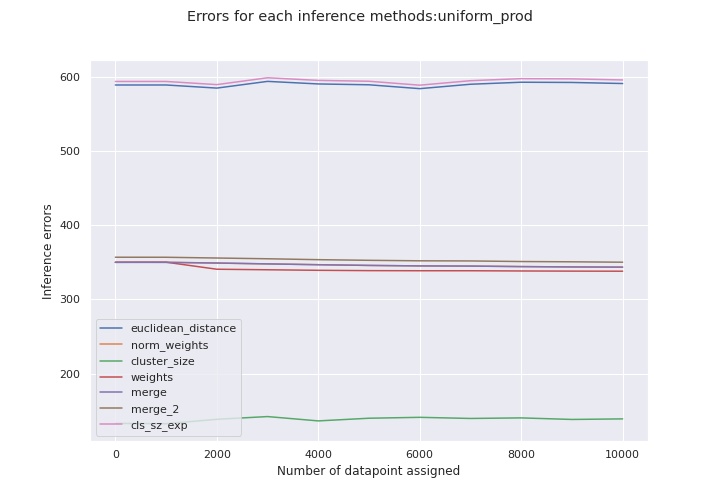}
    \\[\smallskipamount]
    \includegraphics[width=.24\textwidth]{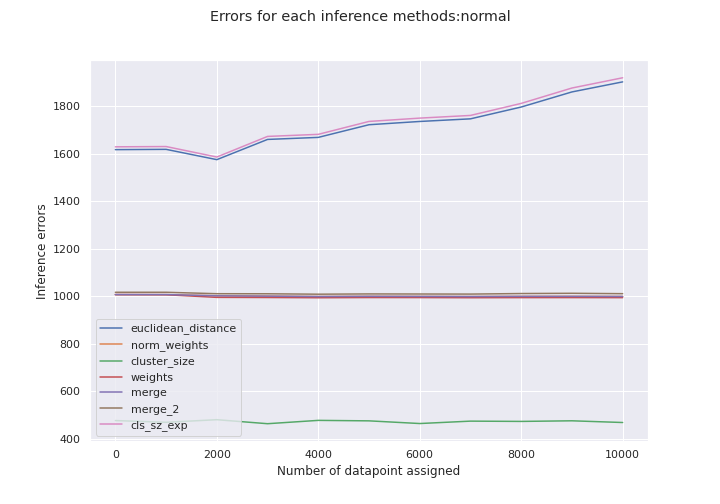}\hfill
    \includegraphics[width=.24\textwidth]{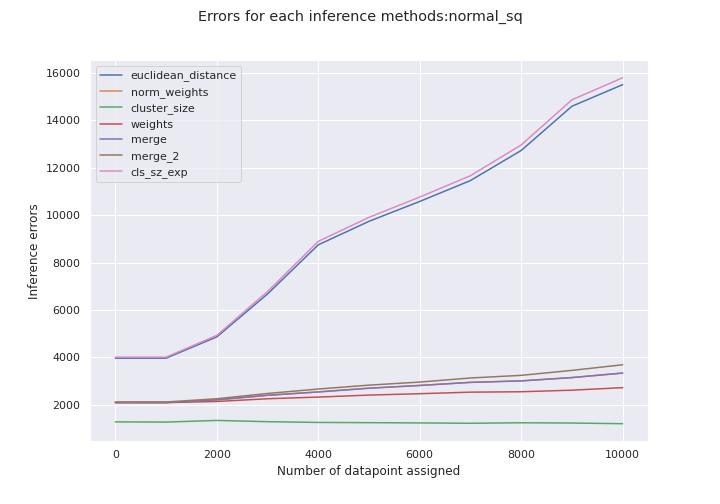}\hfill
    \includegraphics[width=.24\textwidth]{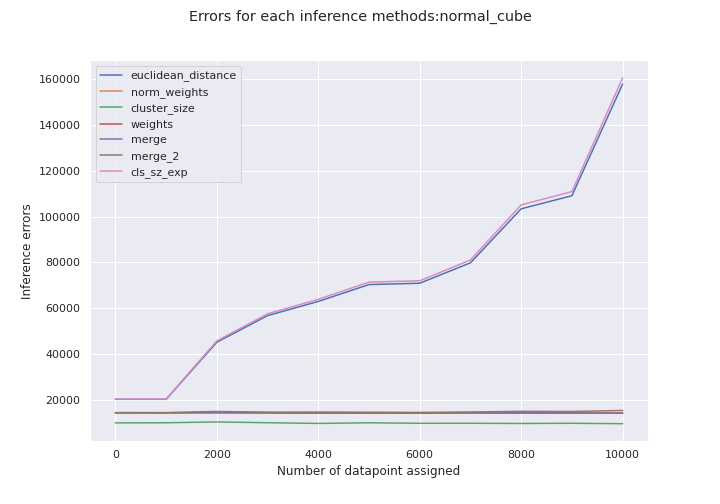}\hfill
    \includegraphics[width=.24\textwidth]{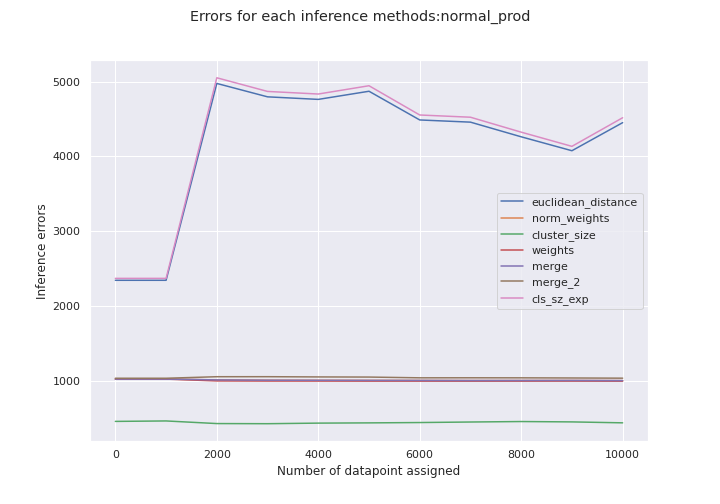}
    \\[\smallskipamount]
    \includegraphics[width=.24\textwidth]{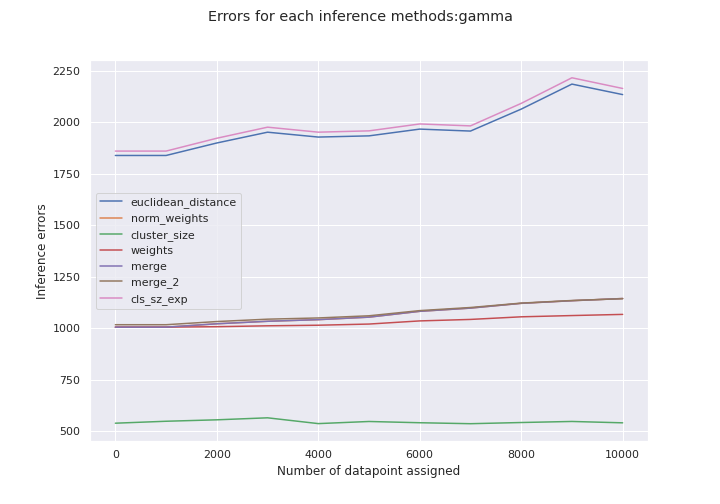}\hfill
    \includegraphics[width=.24\textwidth]{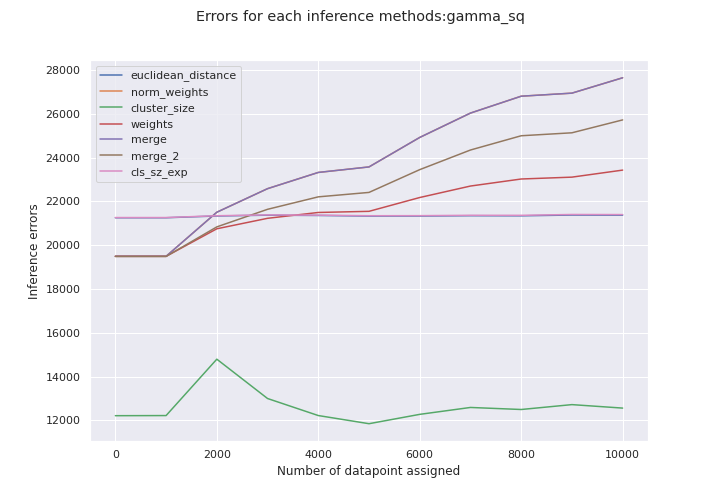}\hfill
    \includegraphics[width=.24\textwidth]{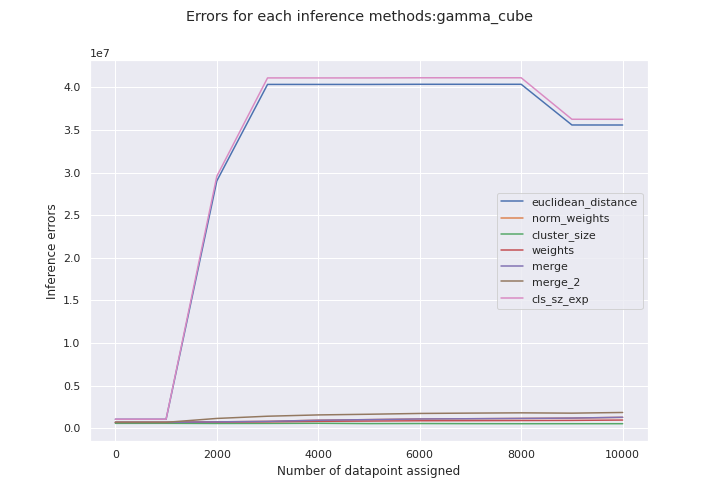}\hfill
    \includegraphics[width=.24\textwidth]{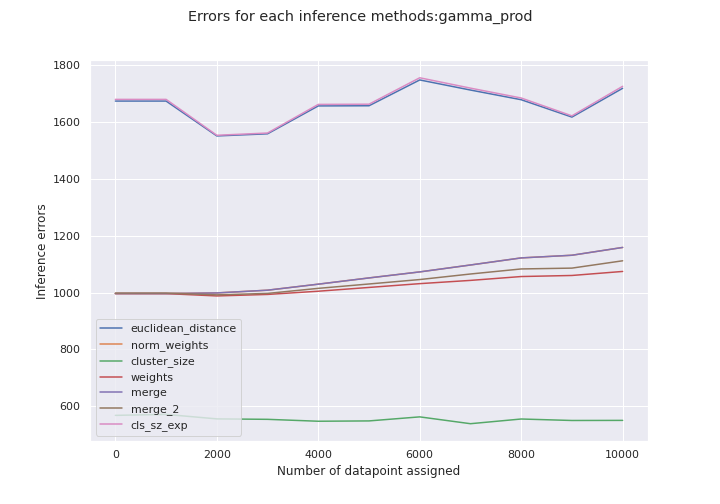}
    \\[\smallskipamount]
    \includegraphics[width=.24\textwidth]{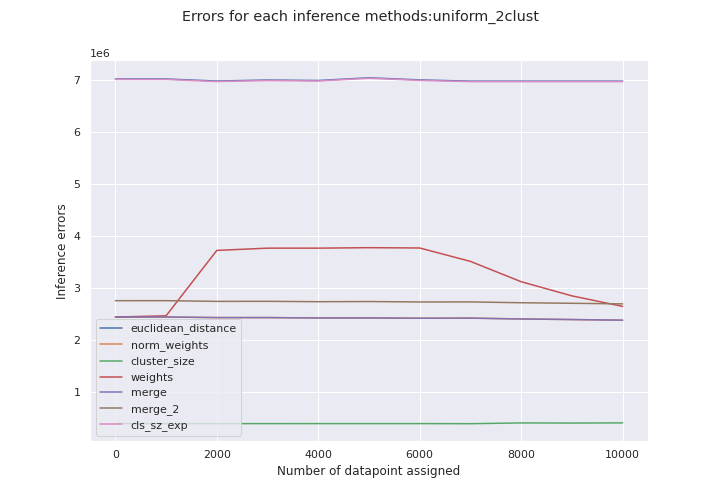}\hfill
    \includegraphics[width=.24\textwidth]{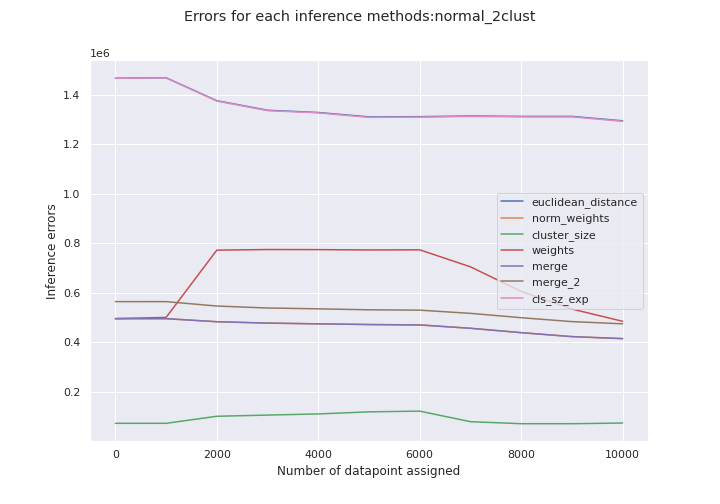}\hfill
    \includegraphics[width=.24\textwidth]{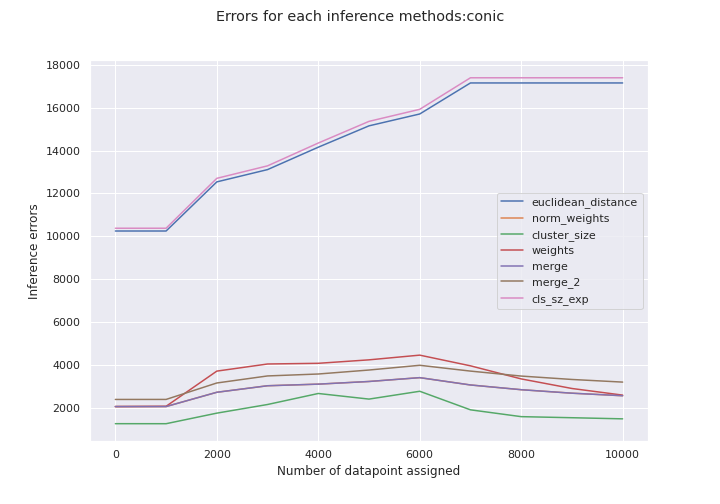}\hfill
    \includegraphics[width=.24\textwidth]{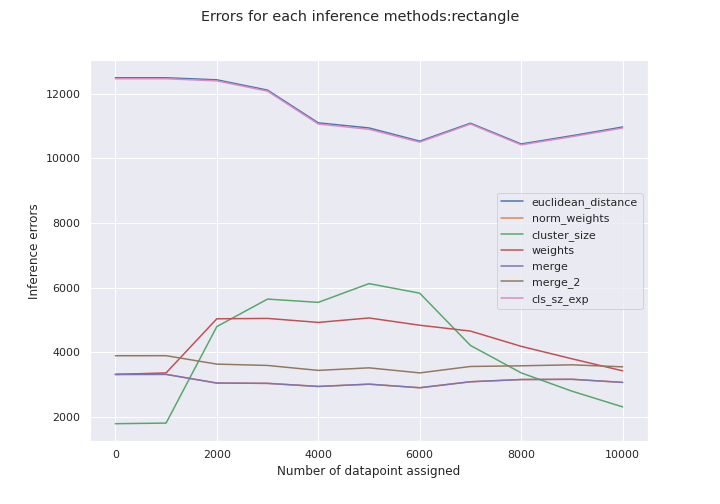} 
    \\[\smallskipamount]
    \includegraphics[width=.24\textwidth]{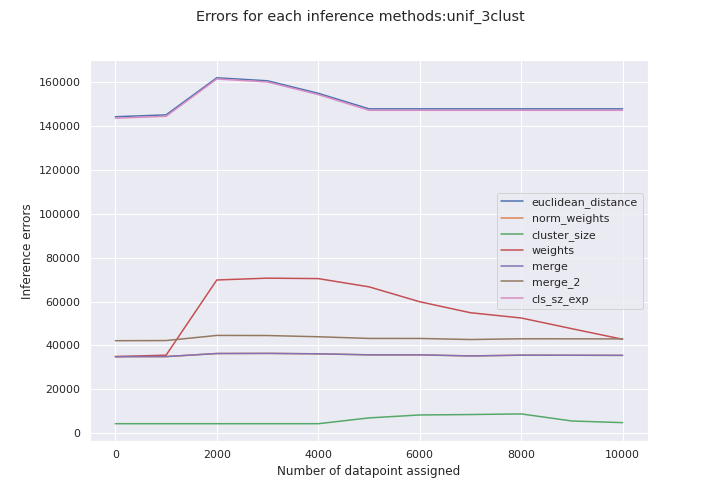}\hfill
    \includegraphics[width=.24\textwidth]{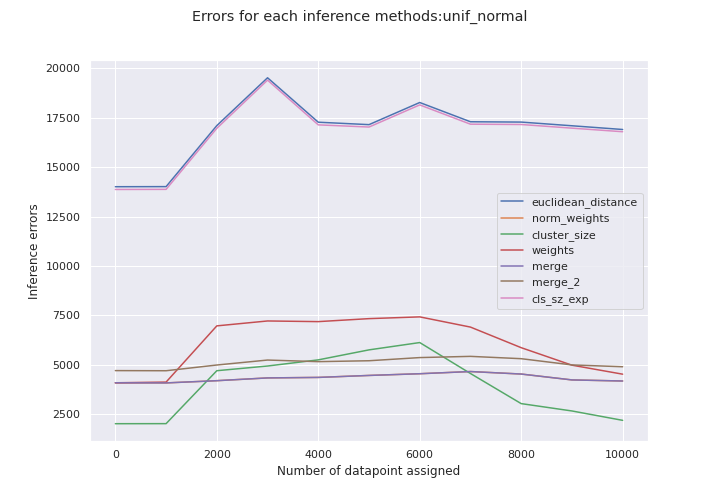}\hfill
    \includegraphics[width=.24\textwidth]{images/unif_3clust.png}\hfill
    \includegraphics[width=.24\textwidth]{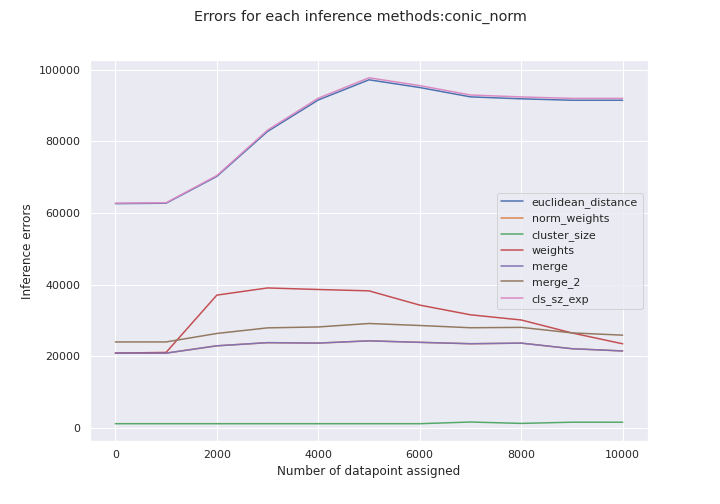}
    \\[\smallskipamount]
    \includegraphics[width=.24\textwidth]{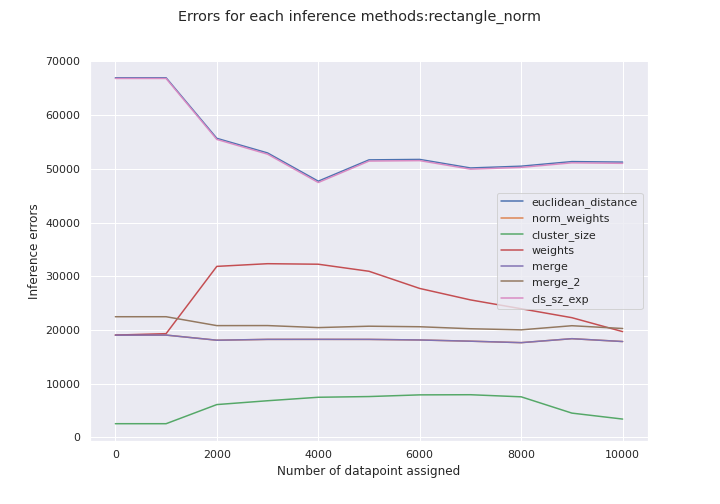}\hfill
    \includegraphics[width=.24\textwidth]{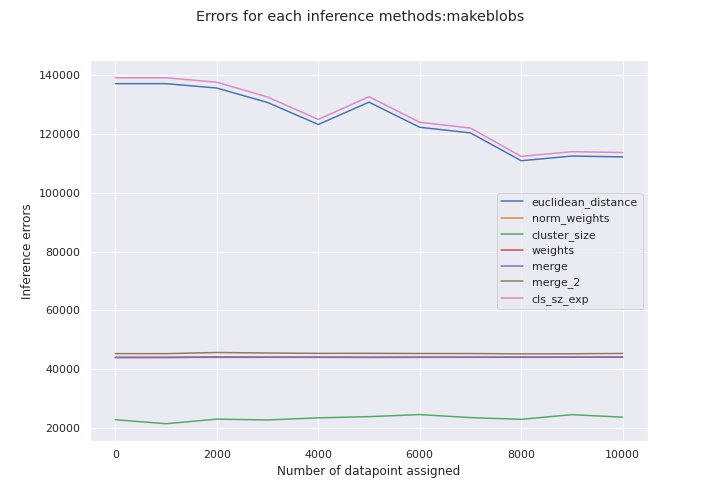}\hfill
    \includegraphics[width=.24\textwidth]{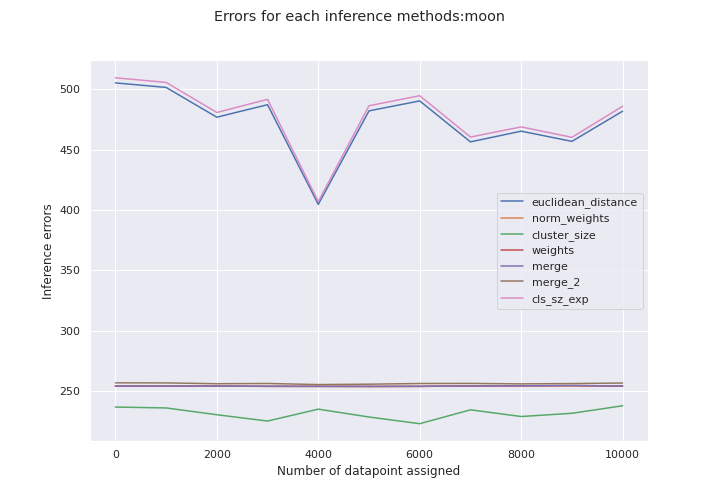}\hfill
    \caption{Plots of errors for each generated data over the individual inference methods}\label{fig:error_data}
\end{figure}

To gain deeper insights into the error propagation, we analyze the errors associated with each inference method by computing the errors for individual data instances. The resulting errors are presented in Figure \ref{fig:error_data}. The curves indicate consistent error levels across all data assignments, albeit with some fluctuations. Notably, the merged normalized weights and cluster size approach exhibited the poorest performance, followed by the Cluster size with exponential weights method. The mentioned methods consistently yielded higher errors compared to the other inference approaches across all data instances.


Finally, our results show that the online balanced k-means model gets better for any chosen hyperparameter, however, the performance of the inference methods remains the same irrespective of the number of data points that have been trained on. Indicating that inference with k-means is not feasible using the proposed inference methods.

\section{Conclusion}

In this research project, we focused on inventing new approaches to making inferences with the k-means algorithm, specifically targeting the prediction of the last component of data points obtained from a distribution of clustered data. Through extensive experimentation and analysis, several key findings have emerged, shedding light on the performance and potential of the proposed methods.

\subsection{Key Findings}

In examining the performance of the developed inference methods, it was observed that a larger number of clusters or partitions tends to yield lower errors. However, it was also found that increasing the number of assigned data points does not significantly improve the inference errors. This suggests that there might be limitations in the performance gains achieved through scaling up the data.

Furthermore, the analysis revealed that reducing losses in the inference process does not significantly impact the overall inference errors. This implies that alternative strategies or modifications to the algorithm might be necessary to improve inference accuracy.

Based on the experimental results and analysis of the data, the best hyperparameters identified for the given dataset were found to be $\alpha$ = 0.6 and $\beta$ = 0.07. It is important to note that the choice of hyperparameters heavily depends on the specific properties of the data under consideration, particularly the number of data points within the dataset, which plays a crucial role in determining the optimal number of clusters. In my experiment, I worked with ten thousand sample points for my training dataset and one thousand for my test dataset. As a result, the best number of clusters is k = 300.

\subsection{Recommendations}

The findings of this research project highlight the need for further investigation and development of specialized inference techniques tailored to better estimate data points derived from multi-clustered data. 

Additionally, it is recommended to explore methods that can yield better results as more data is assigned to the learning algorithm. This can involve investigating algorithms that can adapt and leverage larger datasets without sacrificing inference accuracy. By considering the scalability of the inference methods, we can ensure that they remain effective and efficient in handling increasingly larger datasets.

In conclusion, the research has made strides in advancing the field of inferences with the k-means algorithm. The key findings provide insights into the relationship between cluster sizes, errors, and the impact of data assignment. The identified best hyperparameters offer guidance for researchers working with similar datasets. Our work proved that inference with k-means does not work with the data used and inference method. The recommendations highlight future directions for research, focusing on multi-clustered data estimation and scalability. By pursuing these avenues, we can further enhance the performance and applicability of the k-means algorithm in making accurate inferences.





\paragraph{Acknowledgments}

			This work represents my final thesis at the African Institute of Mathematical Sciences (AIMS) of Ghana. Special gratitude is extended to Matthew Cook for his invaluable insights that led to the development of this thesis. Last but not least, my sincere appreciation to my supervisors, Dr. Alexander Nedergaard and Dr. Prudence Djagba for their relentless support and intellectual assistance in making this work a success.

\newpage

\bibliography{biblio_jmmbiblio_jmm}
\bibliographystyle{plain}
\end{document}